# Procedural Content Generation via Generative Artificial Intelligence

Xinyu MAO[1], Wanli YU[2], Yuya OKAWARA[2], Xueying ZHAN[2],
Kazunori D. YAMADA[2,3] and Michael R. ZIELEWSKI[2,3,*]

[1]*Pratt School of Engineering, Duke University, Durham, NC*
[2]*Graduate School of Information Sciences, Tohoku University, Sendai 980-8579, Japan*
[3]*Unprecedented-scale Data Analytics Center, Tohoku University, Sendai 980-8572, Japan*

The attempt to utilize machine learning in procedural content generation (PCG) has been made in the past. In this survey paper, we investigate how generative artificial intelligence (AI), which saw a significant increase in interest in the mid-2010s, is being used for PCG. We review applications of generative AI for the creation of various types of content, including terrains, items, and even storylines. While generative AI is effective for PCG, building high-performance models requires not only handling customized content and ensuring quality and diversity, but also securing sufficient training data. For PCG research to advance further, addressing these challenges is essential. Thus, we also give special consideration to research that explores innovative generation techniques, model architectures, and approaches suited for limited-data scenarios.

## 1. Introduction

Video game technologies have advanced significantly over time, with one of the most perceptible transitions being the evolution from early pixel-based graphics to modern high-resolution, physics-based rendering. Enabling many of these advancements was a change in the approach to content generation. One key development in this area was procedural content generation (PCG), which refers to the automated creation of game content through algorithmic methods [1].

In the context of PCG, "content" includes, but is not limited to, terrain, characters, items, stories, and music. Developing such content traditionally requires a significant amount of human effort. PCG emerged as an attractive alternative that could reduce development costs and increase content scale through automation. These benefits are particularly appealing to small development teams with limited resources and large scale environments requiring structured yet randomized content. Furthermore, one genre where PCG is especially useful is roguelike games, which often demand the frequent generation of unique content for relatively short game sessions [2].

Early PCG methods employed simple randomization and rule sets to generate content [3]. Over time, they became more sophisticated, especially with the rise of artificial intelligence (AI) in the early 2000s, driven by deep neural networks [4]. While early AI advances focused on computer vision and language, AI has expanded rapidly to many other domains [5]. In the mid-2010s, generative AI in particular drew attention for producing high-quality content without human guidance [6–8], demonstrating the capability of AI to successfully generate text, images, music, videos, and more. Many of the breakthroughs in AI have also transformed PCG, enabling the generation of more complex content in games, including environments, behaviors, and narratives [9].

Providing means to generate content that was previously thought too complex to automatically generate is only one of the motivations of using AI for PCG. Other reasons include addressing player-percieved limitations of non-AI PCG methods. One such limitation is the presence of inconsistent content, which can take different forms depending on the domain. In level generation, for instance, a generated layout may break expected gameplay logic, resulting in a level that cannot be finished. Another limitation is limited content variety, where content that is algorithmically distinct appears identical to human players. To address these issues, there is also growing interest in combining AI with PCG to create more tailored content based on human feedback [10].

Although the term "game" can refer to various formats, including board and card games, this paper focuses on video games. Several surveys on PCG exist [11, 12], including reviews of machine learning applications in PCG [4, 9], and reinforcement learning approaches to formulating PCG problems [13]. However, no prior survey has focused

Received December 11, 2025; Accepted February 3, 2026; J-STAGE Advance published March 10, 2026
This work was supported in part by the Top Global University Project of the Ministry of Education, Culture, Sports, Science and Technology of Japan (MEXT).
*Corresponding author. E-mail: mike.zielewski@tohoku.ac.jp

specifically on generative AI, though there is one addressing the use of Transformers [14] in PCG [15]. This paper fills that gap by reviewing how generative AI is used in PCG, covering past developments and recent trends.

We provide a comprehensive overview, beginning with early AI-based PCG using traditional machine learning, followed by recent applications of generative AI. Finally, we offer insights into the future of game development and gameplay based on our findings.

## 2. Background

### 2.1 The rise of procedural content generation

PCG is the semi-automatic generation of in-game content using algorithms, often operating with minimal input or instructions. Its scope includes terrain, characters, items, stories, music, and other game elements. While typically used to support development, some games integrate PCG as a core design feature.

One main motivation for adopting PCG is cost reduction, which is typically achieved through reducing human labor. Modern games often feature expansive worlds filled with diverse items and characters. Manually creating such content may require hundreds of hours across large teams, making it impractical for smaller developers [3, 12]. PCG enables small teams to build high-quality games comparable to those from large studios.

PCG also supports novel gameplay experiences by enabling content at a scale that would be infeasible to produce and maintain manually. In roguelike games, for instance, content varies with each playthrough, making PCG essential rather than optional. Such games depend on PCG to not only deliver fresh and varied content, but also serve as a form of data compression, reducing storage demands.

For these reasons, and more, PCG has long been used in game development. One of the earliest examples is *Rogue* (1980), a dungeon exploration game where levels, monsters, and items are randomly generated each time. Although *Rogue* relies mainly on randomness, it also uses rule-based constraints to ensure consistency. For example, rooms placed randomly on a map are connected using specific rules. Without such constraints, purely random generation could break game balance or create unplayable layouts.

### 2.2 Procedural content generation and artificial intelligence

AI can broadly be defined as computer-based systems that imitate human intelligence. Under this definition, the randomness-based content generation methods used in *Rogue* and similar roguelike games qualify as a form of basic AI. The inclusion of human-designed rules does not invalidate this interpretation, as such rules typically serve to constrain the space of possible content and guide the generation toward functionality and meaning.

However, as video games have evolved, so too have expectations of PCG, and there is demand for content generation that is more complex and dynamic. In response, interest has shifted from purely rule-based approaches to machine learning methods that leverage data derived from player behavior and experience [9]. This trend has accelerated with the broader rise of interest in AI, leading to the development of PCG methods that incorporate recent technological advances.

A major branch of AI is machine learning, in which modern models are commonly categorized as either generative or discriminative [16]. Generative models learn from real data to synthesize new samples, and are therefore well suited for procedural content generation. Discriminative models, in contrast, focus on modeling the conditional relationship between data and target variables. Although this role may seem to have limited relevance to PCG, discriminative models are integral to the unique training process of generative adversarial networks (GANs), one representative generative model. Other generative models include diffusion models and large language models (LLMs) such as ChatGPT.

Reinforcement learning is another major AI approach, where an agent learns to act in an environment by maximizing cumulative rewards through trial and error [17]. Common algorithms include Q-learning and deep Q-networks. AlphaGo [18], which defeated a world champion in the board game Go, is a landmark example.

Despite its success in gameplay, applications of reinforcement learning to PCG remain relatively rare. Some studies use reinforcement learning to directly generate content, such as stronger card decks [19]. Others employ PCG-generated environments to improve generalization performance of reinforcement learning agents [20]. Prior work has also explored how reinforcement learning could be effectively adapted for PCG tasks [13].

In addition to reinforcement learning and generative models, classical machine learning methods have been widely applied in PCG research [9]. These include support vector machines, decision trees, and sequence models such as Markov chains [21] and long short-term memory networks [22], which treat content generation as a sequence learning problem.

## 3. Targets of Procedural Content Generation

### 3.1 Types of content

Games, as interactive media, have distinct content generation requirements compared to non-interactive media. In particular, most game content must be coherent, as lack of coherence diminishes aesthetic quality [4].

Different content types serve different functions and thus impose varying requirements. Core elements, such as level designs or storylines, must be understandable or playable, with entertainment value as a secondary goal. More relaxed constraints and lower quality is tolerated for less critical assets, like minor character sprites or background textures, enabling trade-offs among various other factors.

## 3.2 Game environments

### 3.2.1 2D game levels

Game levels are core components in genres such as platformers, role-playing games, and roguelikes. In procedural generation research, a level typically refers to a unit of playable scene. Since most studies distinguish 2D level generation from 3D terrain generation, this section focuses on 2D levels only.

A key feature of game levels is their strict functional requirements. Unlike faces or sentences, levels with blocked paths or unreachable goals are unplayable, regardless of visual quality [23]. Thus, ensuring playability is the primary goal, with coherence, diversity, and other factors considered secondary enhancements.

Recent research on generative AI for level generation often builds on GANs, with approaches increasing in complexity to improve quality and control. Irfan *et al.* applied Deep Convolutional GANs to general video game levels [24]. Fontaine *et al.* used Quality-Diversity algorithms to explore the GAN latent space [25]. Torrado *et al.* proposed Conditional Self-Attention GAN that improves long-range consistency and incorporates level features like enemy count as input conditions [23].

### 3.2.2 3D terrain

Terrain, often used in 3D open world games and sandbox games, refers to the topography in virtual environments, typically represented by a heightmap.

Research on terrain modeling have spanned over thirty years, accumulating many classic methods. Fractal Subdivision, one of the oldest and most common procedural terrain generation techniques, generates terrain by recursively increasing the resolution of the terrain and adjusting the height of each new point [26]. The terrains produced by this method typically have a fractal, self-similar structure and details. Other mainstream or efficient procedural terrain generation methods include evolutionary algorithms, erosion simulation, parameter-controlled methods, and user-assisted methods.

In the field of generative AI, Guérin *et al.* present a novel example-based authoring pipeline for generating realistic virtual terrains using conditional GANs (cGANs) [27]. They demonstrate that cGANs offer significant advantages in terrain generation, including enhanced realism, user control, efficiency, learning from real-world examples, and ease of use, making them a superior choice over traditional terrain modeling techniques.

## 3.3 Art and visual assets

### 3.3.1 Sprites

Sprites are graphic elements widely used in game development. Since their appearance in early computer and video games, they have remained essential in representing visual components such as character actions (e.g., walking, running, jumping), background elements, and 2D map tiles. To achieve smooth animation, each action typically consists of multiple frames. For efficient rendering and storage, these frames are often organized into sprite sheets, which combine many small images into a single larger one [28].

As game projects grow in scale, the number of sprites increases, making their management more complex. Adding, removing, or modifying sprites requires manual adjustments, which are both time-consuming and error-prone. This challenge becomes more significant in cases where content must be updated frequently or loaded dynamically during gameplay. Sprite sheets, in particular, complicate this process due to their tightly packed format, which is optimized for performance but difficult to modify flexibly.

In recent years, there has been a growing interest in combining generative networks with sprites to find better methods for improving the efficiency of image generation and processing [29, 30].

### 3.3.2 Motion

In games, human movements such as walking, running, and jumping are typically animated using skeletal animation. This technique models a character as a mesh driven by a hierarchical structure of interconnected bones, separating the model into bones and skin. By manipulating the bones, developers can generate realistic and efficient animations for characters and other objects.

Producing high-quality skeletal animations requires adherence to anatomical and physical principles, as well as the ability to reflect personality and emotion through motion. This makes animation production a technically demanding task that requires both expertise and specialized tools.

Generative networks have recently shown strong potential in this domain [31–33]. These models can synthesize realistic and expressive animation sequences. This not only reduces the manual workload but also improves the realism and emotional quality of animations.

#### 3.3.3 Character face

Facial customization systems are essential in modern games and virtual character platforms. They allow players to design character faces according to their preferences and are commonly used in role-playing games, simulation games, and social environments. Generative networks are increasingly applied to generate and refine facial features based on player input, enhancing engagement [34, 35].

Character creation typically begins with a neutral face. This base model is designed with minimal expression and serves as a consistent foundation for further customization and animation. Developers can then apply various expressions while maintaining visual coherence.

With GANs, it is possible to separate identity and expression parameters. This enables systems to preserve unique facial features while generating a wide range of emotional expressions. Such capability is especially valuable in customization systems, as it supports expressive yet personalized character creation.

### 3.4 Narrative components

Narrative components are essential to completing a game, providing players with motivation and contextual grounding.

Game narratives consist of story and discourse [36]. The story includes plot (structured events) and space (characters and settings), while discourse determines how the story is presented.

A notable example of generative AI applied to game narratives is the work of Park *et al.* [37]. They introduced generative agents with memory and planning capabilities, enabling emergent, dynamic narratives shaped by player interactions. These agents inhabit the narrative space through distinct personalities and behaviors. Their interactions generate evolving plots, while the player's perception of these interactions forms the discourse. This approach enables unique, responsive stories that change with each playthrough, enhancing immersion and replayability. By leveraging generative AI in this way, developers can create dynamic worlds where each session yields a different narrative experience.

### 3.5 Music and sound effects

Although not essential to gameplay mechanics, music and sound effects play a key role in enhancing immersion.

Traditional sound synthesis methods face challenges in quality, diversity, efficiency, and usability. Sampling-based approaches require pre-recorded materials and manual mixing, which is time-consuming and labor-intensive [38]. Procedural synthesis generates sound in real time using mathematical models but often produces low-quality results that fall short of recorded sound effects [39]. These methods also struggle to generate enough variation, leading to auditory fatigue and reduced immersion [40].

Recent deep learning approaches aim to overcome these limitations. For example, Andreu and Aylagas propose a method to generate controllable variations of sound effects for use in game sound design [41]. They utilize WaveFlow, a generative flow model operating directly on raw audio, originally developed for speech synthesis.

## 4. Methods for Generating Content

### 4.1 Representative generative methods

This survey focuses on three representative generative methods for PCG: GANs, diffusion models, and Transformers. These methods were selected because they represent the dominant and most widely adopted paradigms of modern generative AI in PCG research [5, 9].

Together, they cover a broad range of generation mechanisms and application domains. GANs have been extensively used for visual and spatial content generation in games and form a large body of early generative AI based PCG work. Diffusion models have recently gained attention due to their stable training and high quality outputs, particularly in motion, material, and level generation. Transformers, especially LLMs, enable controllable and interactive generation through sequence modeling and natural language interfaces, which is increasingly important for game development workflows.

Other generative models exist, but their application to PCG has so far been more limited or often appears as supporting components within these approaches, making GANs, diffusion models, and Transformers a practical focus for a concise review.

### 4.2 Generative adversarial networks

GANs are a kind of generative model based on game theory [42]. They train a generator to produce fake data and a discriminator to distinguish it from real data. This adversarial training improves both networks over time.

In 2018, Giacomello *et al.* applied GANs to PCG for the game *DOOM* [43]. They trained two models on over 1000 human-designed levels: an unconditional GAN using level images and a cGAN using additional topological features. Each level was processed into six image types (e.g., walls, floor height) and various numerical and categorical attributes (e.g., number of rooms, equivalent diameter). The cGAN generated levels more structurally similar to human designs. Evaluation used custom metrics inspired by indoor map assessment.

Volz *et al.* applied GANs to level generation for *Super Mario Bros* [44]. They trained a GAN on a single level using one-hot encoded tile types and showed that it could generate playable levels resembling human designs. They also introduced latent variable evolution (LVE) to search the GAN's latent space using Covariance Matrix Adaptation Evolution Strategy. In one approach, the fitness function depended on the presence of specific tile types, allowing control over content and difficulty. Another approach used an agent-based fitness function based on gameplay behavior. Results showed that both methods effectively adjusted level difficulty. This work demonstrated that GANs can learn level structures even with limited data and that LVE can guide generation toward desired properties.

Inspired by Volz *et al.* [44], Irfan *et al.* [24] applied a similar GAN-based approach using image data instead of tile types. Their training data was generated by a random level generator and included three games: *Freeway*, *Zelda*, and *Colourscape*, selected for their varying difficulty levels. Initially, the GAN failed to produce valid levels but gradually improved. After evaluation with game-playing agents, fewer than 20% of generated levels were unplayable. This study showed that image-based training is viable for level generation and highlighted the flexibility of GANs across different game types.

A common challenge in using GANs for PCG is the need for domain-specific training data. Wulff-Jensen *et al.* [45] addressed this by generating 3D landscapes using GANs trained on real-world digital elevation data from Norway and the Alps. Instead of relying on scarce hand-crafted landscapes, this approach allowed the model to learn terrain features such as hills and ravines from real locations. The generated landscapes were found to be realistic, relatable, and traversable, demonstrating that real-world data can support the generation of high-quality 3D game environments.

Park *et al.* applied GANs to level generation in educational games, where diverse difficulty levels are needed to support varying player competencies [46]. To reduce the cost of manual level creation, they proposed a multistep GAN framework. The first GAN generated novel but potentially unsolvable levels from human-designed samples. The second GAN was trained only on solvable levels filtered from the first step. Evaluations showed that this two-stage approach effectively balanced novelty and solvability, demonstrating a way to overcome limitations of single-stage GANs in educational content generation.

Panagiotou *et al.* proposed a sequential GAN-based approach to generate 3D terrains [47, 48]. They first trained a Progressively Growing GAN [49] to generate high-quality satellite-style overhead images. Then, a cGAN was trained to produce Digital Elevation Models conditioned on the generated images, ensuring that the resulting 3D terrains appeared natural. Their method showed that combining multiple GANs can produce visually coherent and believable 3D environments.

Hald *et al.* compared a standard GAN and a cGAN for generating levels in a tile-matching game [50]. The game includes structural, non-interactive, and interactive tile types. The cGAN was conditioned on level shape and tile distribution, with the same information also provided to the discriminator. Models were evaluated based on shape, tile distribution, placement, and combination. Both models performed well in shape and tile combination, while the cGAN outperformed the standard GAN in overall similarity to the training data. This study highlighted the advantage of incorporating additional input information in GAN-based level generation.

Coutinho and Chaimowicz proposed a Pix2Pix-based model to translate pixel art character sprites from one pose to another [51]. Unlike bone-based approaches, their method does not require skeletal data and supports diverse character shapes. The generator uses a U-net to convert $64 \times 64$ RGBA images into target poses, while a PatchGAN discriminator evaluates local image quality [52]. Experiments showed the model could transfer colors and shapes to some extent, though issues like color inaccuracy, noise, and blurring remained. Nonetheless, the model demonstrated a degree of generalization for pose translation tasks.

In 2020, Shi *et al.* proposed PokerFace-GAN, a generative framework for automatic game character creation based on facial similarity [53]. It consists of three components: a predictor for facial parameter estimation, a differentiable renderer that converts parameters into 3D characters, and a discriminator trained on expressionless faces to ensure neutrality. The predictor uses LightCNN-29v2 [54] and ResNet-50 [55] to extract facial features, which are translated into identity, expression, and pose parameters. The system was trained on the CelebA dataset [56] and successfully generated 3D characters closely resembling input photos, demonstrating the potential of combining adversarial learning with differentiable rendering for procedural character generation.

Donahue *et al.* introduced WaveGAN and SpecGAN, GAN-based models for raw audio and spectrogram generation, respectively [57]. To address audio-specific artifacts, they proposed techniques such as phase shuffling. Experiments on datasets like spoken digits and environmental sounds showed that both models could produce coherent audio. While SpecGAN achieved higher inception scores, human evaluators preferred WaveGAN for its sound quality and speaker diversity. These results highlight the potential of GANs for unsupervised audio synthesis and creative sound design.

### 4.3 Diffusion models

A diffusion model is a Markov chain that gradually adds noise to data and learns to reverse the process to generate structured outputs. It uses neural networks to model Gaussian transitions and produces samples matching the original distribution [58].

Zhang *et al.* proposed MotionDiffuse, a diffusion-based model that generates human motion from text [59]. It offers probabilistic sampling for greater diversity, fine-grained control from text input, and coordination of body parts over time. Results show that MotionDiffuse outperforms existing methods and produces natural-looking motion, demonstrating the strength of diffusion models in complex content generation.

To address the challenge of long human motion sequences, Zhang *et al.* later proposed aligning the motion timeline with the diffusion process and introducing a buffer of temporally-varying noise [60]. This allows smooth transitions and realistic motion over time. They also support guided generation by replacing frames with target poses, enabling transitions toward specified movements. Evaluation confirms the effectiveness of this method in generating long-term natural motion.

Tevet *et al.* introduced the Motion Diffusion Model (MDM), a general framework for human motion generation that handles multiple conditions, such as text and action classes [61]. MDM achieves state-of-the-art results while being more efficient than previous models. A key contribution is the integration of geometric loss functions (e.g., positions, foot contact, velocities), which enforce physical realism and reduce artifacts in generated motion.

For material generation in graphics, Vecchio *et al.* proposed MatFuse, a diffusion model for generating Spatially-Varying Bidirectional Reflectance Distribution Function maps [62]. MatFuse supports multiple conditioning types, including text, palettes, images, and sketches, applied at both global and local levels. It achieves high controllability and outperforms prior generative methods in evaluations.

Dai *et al.* applied unconditional diffusion models to level generation in *Super Mario Bros* and *Minecraft* [63]. They replaced one-hot vectors with dense word embeddings for pieces, improving scalability and training. To reduce overfitting on single examples, they used residual fully convolutional networks with cropped inputs. The method generated levels with finer details and fewer artifacts compared to other single-sample generative approaches.

### 4.4 Transformers

The Transformer is a neural architecture based solely on attention mechanisms, without recurrence or convolutions [14]. It has revolutionized fields like natural language processing and content generation, especially for sequential data with long-range dependencies.

He *et al.* proposed an AI system for video generation from text queries [64]. The system searches for suitable scenes and motions and composes them to follow a given plot. It combines technologies such as LLMs, Markov chains, and diffusion models. While not designed specifically for games, the method is adaptable to game content generation.

Sudhakaran *et al.* proposed MarioGPT, a Transformer-based method for generating levels in *Super Mario Bros* [65]. Unlike GANs, which require costly latent space searches, MarioGPT enables controllable generation via natural language prompts. Most generated levels are diverse and playable, and the method requires no additional model components beyond a pre-trained LLM, showcasing the potential of LLMs in PCG.

Zhang *et al.* proposed a multimodal approach for human motion generation using a Transformer-based LLM [66]. They incorporate pose data as discrete codes learned via a vector quantized-variational autoencoder [67]. Results show that combining pose and text inputs significantly improves motion quality, especially when using multiple pose points across a sequence. The study highlights the value of multimodal data and suggests further gains may be possible using other modalities like audio.

## 5. Discussion

### 5.1 Challenges in procedural content generation for generative artificial intelligence

In PCG, generative networks face notable challenges in producing visual assets. A major issue is limited diversity [68, 69]. Models often struggle to generate assets with varied styles or types, which results in repetitive or uninspired outputs. Creating complex scenes or detailed characters is also difficult, especially when designing large-scale game worlds. Current models may fail to produce realistic or aesthetically pleasing visuals, and noise artifacts can further reduce quality.

Similar problems occur in level generation. Datasets are usually small and game-specific, which limits generalization. Effective level generation requires understanding both spatial structures and player behavior. This includes designing levels with varying difficulty and skill requirements. However, neural models often overfit to the training data and fail to learn general gameplay logic.

Validation of generated content is another challenge. While visual flaws such as mismatched tiles or colors may be acceptable, functional flaws can make the game unplayable. Examples include blocked paths or unreachable goals. Metrics offer a convenient way to evaluate quality automatically, but they often miss important details. Game-playing agents, such as pathfinding algorithms, are more effective for checking playability, although they are computationally expensive and often provide only binary outcomes [70, 71].

The computational demands of generative AI also present challenges for interactive applications. Deep neural network-based models require GPU resources, which may compete with rendering tasks during gameplay, potentially reducing graphics quality. Real-time content generation further increases the load, and devices with limited hardware may struggle to maintain a smooth user experience.

### 5.2 Current approaches to procedural content generation with generative artificial intelligence

The reviewed works employ various generation techniques, with clear trends over time. GANs were initially popular due to their breakthrough in generative modeling [7]. Later, diffusion models and transformers gained traction by offering more stable training, better convergence, and higher-quality outputs [58]. Transformers also allow natural language input, making them more accessible [72]. These shifts reflect how advances in neural network technologies quickly redefine best practices, requiring flexibility from developers.

Many studies combine multiple techniques to improve results. For instance, a standard GAN may fail to consistently produce desired outputs. Using a cGAN allows some control, but still does not guarantee specific features. Searching the latent space with optimization, such as in [44], can guide generation toward target properties. This shows that combining models and algorithms is crucial for high-quality generation.

Generative AI typically uses supervised or unsupervised learning, in contrast to reinforcement learning. Generative methods learn from large datasets, while reinforcement learning relies on predefined rules and reward signals. In PCG, generative AI is suited to domains with abundant data. When data is scarce, reinforcement learning may be more appropriate. This distinction provides a useful guideline for method selection.

### 5.3 Potential of generative artificial intelligence as game mechanics

PCG not only improves development efficiency but can also serve as a game mechanic to enhance player experience or create new forms of gameplay.

One area in which generative AI is particularly powerful is narrative creation. Recent advances in natural language processing allow non-player characters (NPCs) to speak and act according to distinct personalities. For example, in *Treacherous Waters Online*, every NPC in the main town uses a ChatGPT-based model to engage in real-time conversation. In *The Portopia Serial Murder Case*, players solve mysteries by freely conversing with characters, replacing fixed dialogue options. These innovations mark a new direction in narrative interaction and merit further exploration.

User-Generated Content (UGC) refers to content created by users rather than developers [73]. UGC has gained popularity in games like *Eggy Party*. Generative AI can support UGC editors by generating textures, images, or 3D models on demand. Unlike traditional editors that rely on fixed assets, generative tools reduce effort and allow for more personalized and stylized content creation.

### 5.4 Future developments for generative artificial intelligence in procedural content generation

#### 5.4.1 Improving quality, usability, and data efficiency

With the growth of generative AI in PCG, future work will likely focus on improving content quality and usability. While higher quality is a natural goal, it is not easily achieved and often requires combining multiple techniques. Usability is equally important, as models that are hard to train or use will not gain traction in practice.

Combining technologies is already common. For example, search algorithms are used to explore GAN latent spaces, and multimodal models incorporate different data types to influence output [74, 75]. Such approaches help generate content with specific characteristics, and we expect future work to build on this trend by integrating techniques from various fields.

Data scale also affects content quality. Although limited training data is a known issue in PCG, it is not unique to games. Other fields, like medicine, use data augmentation to address this [76, 77]. However, such techniques are rarely applied in games, partly because simple transformations (like vertical flips) may distort game logic. Since data augmentation is effective elsewhere, more research is needed to explore suitable transformations for game data.

Ease of use is another key factor. Complex models may reduce the time savings that PCG promises. Transformers and LLMs offer a solution by allowing developers to generate content via natural language prompts. These models are both powerful and easy to use. We expect future systems to incorporate text input and feedback mechanisms for more accessible PCG workflows.

#### 5.4.2 Procedural content generation beyond static content

Recent work has extended PCG beyond static assets toward interactive models that explicitly capture environment dynamics. In this direction, PCG is treated as the generation of executable environments whose states evolve in response to agent actions, thereby blurring the boundary between content generation and game engines.

Early examples include World Models [78] and GameGAN [79], which learn latent representations of environment dynamics from interaction or gameplay data and enable the simulation of state transitions conditioned on actions. World Models demonstrate that agents can be trained within internally generated environments that approximate real dynamics, while GameGAN shows that visually realistic and temporally coherent game environments can be learned directly from gameplay videos.

More recent systems such as Genie [80] and GameNGen [81] further advance this idea by learning more general interactive environments or neural game engines. Genie focuses on large-scale, action-conditioned environment generation from unlabeled video data, whereas GameNGen emphasizes real-time neural simulation, effectively functioning as a learned game engine without hand-crafted dynamics.

Despite the success of these interactive models, many rely on large-scale datasets or complex training pipelines. For example, in GameNGen, a reinforcement learning agent must first be trained to play a game in order to record data used to train the diffusion model acting as the game engine [81]. Furthermore, these models also use image-based representations of data, which make reuse and targeted modification challenging. Consequently, the direct use of interactive world models in PCG remains limited, particularly when control is critical or when used in resource-constrained development environments.

A closely related perspective is PCG for AI, where environments are generated to support the training of robust learning agents. Unsupervised Environment Design [82] formulates environment generation as an adaptive process that exposes agent weaknesses, while Prioritized Level Replay [83] and ACCEL [84] refine this idea by prioritizing challenging environments and co-evolving agents and environments. In these frameworks, PCG is not an end goal but a mechanism for shaping the learning process itself.

Together, these directions suggest that future PCG research will increasingly emphasize interactive, adaptive, and learning-centered generation, especially as current limitations are overcome. Such a shift positions generative AI as a core component of dynamic simulation and training environments rather than solely a tool for static content creation.

## 6. Conclusion

This paper explores the application of generative AI to PCG, focusing on three primary methods: GANs, Transformers, and diffusion models. The application of these generative models in various PCG scenarios is summarized, offering insights for game developers and researchers.

GANs demonstrate a wide range of applications in generating high-quality game environments, characters, and sounds. However, they continue to face challenges with result instability and noise, necessitating further optimization. Transformers excel in handling long sequential data due to their self-attention mechanism, making them particularly suitable for tasks such as movie production, game level generation, and human action generation. Their ability to produce coherent and logical content is a significant advantage in these contexts. Diffusion modeling, an emerging technique that generates high-quality data through gradual noise addition, shows remarkable capabilities in human action generation, material creation, and game level design. Its potential in PCG applications is becoming increasingly evident.

Generative AI provides diverse methods for PCG, showing potential and research gaps for researchers or game developers to dig deeper into. Future research should focus on not only improving quality through model architecture improvements, but also on other considerations necessary for realistic applications of generative AI in PCG, such as content validation and balancing the demand for computing resources and the user experience.

## Author Contributions

All authors conducted the survey, contributed to manuscript writing, and approved the final version of the manuscript.

## Acknowledgments

The authors have used generative AI technologies for proofreading and translating. Specifically, ChatGPT 3.5 and Claude 2.0 were utilized.